# Robust LogitBoost and Adaptive Base Class (ABC) LogitBoost

**Ping Li**
Department of Statistical Science
Faculty of Computing and Information Science
Cornell University
Ithaca, NY 14853
pingli@cornell.edu

## Abstract

**Logitboost** is an influential boosting algorithm for classification. In this paper, we develop **robust logitboost** to provide an explicit formulation of tree-split criterion for building weak learners (regression trees) for *logitboost*. This formulation leads to a numerically stable implementation of *logitboost*. We then propose **abc-logitboost** for multi-class classification, by combining *robust logitboost* with the prior work of *abc-boost*. Previously, *abc-boost* was implemented as *abc-mart* using the *mart* algorithm.

Our extensive experiments on multi-class classification compare four algorithms: *mart*, *abc-mart*, *(robust) logitboost*, and *abc-logitboost*, and demonstrate the superiority of *abc-logitboost*. Comparisons with other learning methods including SVM and **deep learning** are also available through prior publications.

## 1 Introduction

Boosting [14, 5, 6, 1, 15, 8, 13, 7, 4] has been successful in machine learning and industry practice. This study revisits *logitboost* [8], focusing on multi-class classification.

We denote a training dataset by $\{y_i, \mathbf{x}_i\}_{i=1}^N$, where $N$ is the number of feature vectors (samples), $\mathbf{x}_i$ is the $i$th feature vector, and $y_i \in \{0, 1, 2, ..., K-1\}$ is the $i$th class label, where $K \geq 3$ in multi-class classification.

Both *logitboost* [8] and *mart* (multiple additive regression trees) [7] can be viewed as generalizations to the classical logistic regression, which models class probabilities $p_{i,k}$ as

$$p_{i,k} = \mathbf{Pr}\left(y_i = k | \mathbf{x}_i\right) = \frac{e^{F_{i,k}(\mathbf{x_i})}}{\sum_{s=0}^{K-1} e^{F_{i,s}(\mathbf{x_i})}}. \qquad (1)$$

While logistic regression simply assumes $F_{i,k}(\mathbf{x}_i) = \beta_k^\mathrm{T} \mathbf{x}_i$, *Logitboost* and *mart* adopt the flexible "additive model," which is a function of $M$ terms:

$$F^{(M)}(\mathbf{x}) = \sum_{m=1}^M \rho_m h(\mathbf{x}; \mathbf{a}_m), \qquad (2)$$

where $h(\mathbf{x}; \mathbf{a}_m)$, the base (weak) learner, is typically a regression tree. The parameters, $\rho_m$ and $\mathbf{a}_m$, are learned from the data, by maximum likelihood, which is equivalent to minimizing the *negative log-likelihood loss*

$$L = \sum_{i=1}^N L_i, \qquad L_i = -\sum_{k=0}^{K-1} r_{i,k} \log p_{i,k} \qquad (3)$$

where $r_{i,k} = 1$ if $y_i = k$ and $r_{i,k} = 0$ otherwise.

For identifiability, $\sum_{k=0}^{K-1} F_{i,k} = 0$, i.e., the **sum-to-zero** constraint, is usually adopted [8, 7, 17, 11, 16, 19, 18].

### 1.1 Logitboost

As described in Alg. 1, [8] builds the additive model (2) by a greedy stage-wise procedure, using a second-order (diagonal) approximation, which requires knowing the first two derivatives of the loss function (3) with respective to the function values $F_{i,k}$. [8] obtained:

$$\frac{\partial L_i}{\partial F_{i,k}} = -\left(r_{i,k} - p_{i,k}\right), \qquad \frac{\partial^2 L_i}{\partial F_{i,k}^2} = p_{i,k}\left(1 - p_{i,k}\right). \qquad (4)$$

While [8] assumed the sum-to-zero constraint, they showed (4) by conditioning on a "base class" and noticed the resultant derivatives were independent of the choice of the base.

---

**Algorithm 1** Logitboost [8, Alg. 6]. $\nu$ is the shrinkage.

0: $r_{i,k} = 1$, if $y_i = k$, $r_{i,k} = 0$ otherwise.
1: $F_{i,k} = 0$, $p_{i,k} = \frac{1}{K}$, $k = 0$ to $K-1$, $i = 1$ to $N$
2: For $m = 1$ to $M$ Do
3:     For $k = 0$ to $K-1$, Do
4:         Compute $w_{i,k} = p_{i,k}\left(1 - p_{i,k}\right)$.
5:         Compute $z_{i,k} = \frac{r_{i,k} - p_{i,k}}{p_{i,k}\left(1 - p_{i,k}\right)}$.
6:         Fit the function $f_{i,k}$ by a weighted least-square of $z_{i,k}$
:         to $\mathbf{x}_i$ with weights $w_{i,k}$.
7:         $F_{i,k} = F_{i,k} + \nu \frac{K-1}{K}\left(f_{i,k} - \frac{1}{K}\sum_{k=0}^{K-1} f_{i,k}\right)$
8:     End
9:     $p_{i,k} = \exp(F_{i,k}) / \sum_{s=0}^{K-1} \exp(F_{i,s})$
10: End

---

At each stage, *logitboost* fits an individual regression function separately for each class. This diagonal approximation appears to be a must if the base learner is implemented using regression trees. For industry applications, using trees as the weak learner appears to be the standard practice.

## 1.2 Adaptive Base Class Boost

[12] derived the derivatives of (3) under the sum-to-zero constraint. Without loss of generality, we can assume that class 0 is the base class. For any $k \neq 0$,

$$\frac{\partial L_i}{\partial F_{i,k}} = (r_{i,0} - p_{i,0}) - (r_{i,k} - p_{i,k}), \tag{5}$$

$$\frac{\partial^2 L_i}{\partial F_{i,k}^2} = p_{i,0}(1 - p_{i,0}) + p_{i,k}(1 - p_{i,k}) + 2p_{i,0}p_{i,k}. \tag{6}$$

The base class must be identified at each boosting iteration during training. [12] suggested an exhaustive procedure to adaptively find the best base class to minimize the training loss (3) at each iteration. [12] combined the idea of *abc-boost* with *mart*, to develop **abc-mart**, which achieved good performance in multi-class classification.

It was believed that *logitboost* could be numerically unstable [8, 7, 9, 3]. In this paper, we provide an explicit formulation for tree construction to demonstrate that *logitboost* is actually stable. We name this construction **robust logitboost**. We then combine the idea of *robust logitboost* with *abc-boost* to develop **abc-logitboost**, for multi-class classification, which often considerably improves *abc-mart*.

## 2 Robust Logitboost

In practice, tree is the default weak learner. The next subsection presents the tree-split criterion of *robust logitboost*.

### 2.1 Tree-Split Criterion Using 2nd-order Information

Consider $N$ weights $w_i$, and $N$ response values $z_i$, $i = 1$ to $N$, which are assumed to be ordered according to the sorted order of the corresponding feature values. The tree-split procedure is to find the index $s$, $1 \leq s < N$, such that the weighted square error (SE) is reduced the most if split at $s$. That is, we seek the $s$ to maximize

$$Gain(s) = SE_T - (SE_L + SE_R)$$
$$= \sum_{i=1}^{N}(z_i - \bar{z})^2 w_i - \left[\sum_{i=1}^{s}(z_i - \bar{z}_L)^2 w_i + \sum_{i=s+1}^{N}(z_i - \bar{z}_R)^2 w_i\right]$$

where

$$\bar{z} = \frac{\sum_{i=1}^{N} z_i w_i}{\sum_{i=1}^{N} w_i}, \quad \bar{z}_L = \frac{\sum_{i=1}^{s} z_i w_i}{\sum_{i=1}^{s} w_i}, \quad \bar{z}_R = \frac{\sum_{i=s+1}^{N} z_i w_i}{\sum_{i=s+1}^{N} w_i}.$$

We can simplify the expression for $Gain(s)$ to be:

$$Gain(s) = \sum_{i=1}^{N}(z_i^2 + \bar{z}^2 - 2\bar{z}z_i)w_i$$
$$- \sum_{i=1}^{s}(z_i^2 + \bar{z}_L^2 - 2\bar{z}_L z_i)w_i - \sum_{i=s+1}^{N}(z_i^2 + \bar{z}_R^2 - 2\bar{z}_R z_i)w_i$$
$$= \sum_{i=1}^{N}(\bar{z}^2 - 2\bar{z}z_i)w_i - \sum_{i=1}^{s}(\bar{z}_L^2 - 2\bar{z}_L z_i)w_i - \sum_{i=s+1}^{N}(\bar{z}_R^2 - 2\bar{z}_R z_i)w_i$$
$$= \left[\bar{z}^2 \sum_{i=1}^{N} w_i - 2\bar{z} \sum_{i=1}^{N} z_i w_i\right]$$
$$- \left[\bar{z}_L^2 \sum_{i=1}^{s} w_i - 2\bar{z}_L \sum_{i=1}^{s} z_i w_i\right] - \left[\bar{z}_R^2 \sum_{i=s+1}^{N} w_i - 2\bar{z}_R \sum_{i=s+1}^{N} z_i w_i\right]$$

$$Gain(s) = \left[-\bar{z}\sum_{i=1}^{N} z_i w_i\right] - \left[-\bar{z}_L \sum_{i=1}^{s} z_i w_i\right] - \left[-\bar{z}_R \sum_{i=s+1}^{N} z_i w_i\right]$$
$$= \frac{\left[\sum_{i=1}^{s} z_i w_i\right]^2}{\sum_{i=1}^{s} w_i} + \frac{\left[\sum_{i=s+1}^{N} z_i w_i\right]^2}{\sum_{i=s+1}^{N} w_i} - \frac{\left[\sum_{i=1}^{N} z_i w_i\right]^2}{\sum_{i=1}^{N} w_i}$$

Plugging in $w_i = p_{i,k}(1 - p_{i,k})$, $z_i = \frac{r_{i,k} - p_{i,k}}{p_{i,k}(1 - p_{i,k})}$ yields

$$Gain(s) = \frac{\left[\sum_{i=1}^{s}(r_{i,k} - p_{i,k})\right]^2}{\sum_{i=1}^{s} p_{i,k}(1 - p_{i,k})} + \frac{\left[\sum_{i=s+1}^{N}(r_{i,k} - p_{i,k})\right]^2}{\sum_{i=s+1}^{N} p_{i,k}(1 - p_{i,k})}$$
$$- \frac{\left[\sum_{i=1}^{N}(r_{i,k} - p_{i,k})\right]^2}{\sum_{i=1}^{N} p_{i,k}(1 - p_{i,k})}. \tag{7}$$

There are at least two ways to see why the criterion given by (7) is numerically stable. First of all, the computations involve $\sum p_{i,k}(1 - p_{i,k})$ as a group. It is much less likely that $p_{i,k}(1 - p_{i,k}) \approx 0$ for all $i$'s in the region. Secondly, if indeed that $p_{i,k}(1 - p_{i,k}) \to 0$ for all $i$'s in this region, it means the model is fitted perfectly, i.e., $p_{i,k} \to r_{i,k}$. In other words, (e.g.,) $\left[\sum_{i=1}^{N}(r_{i,k} - p_{i,k})\right]^2$ in (7) also approaches zero at the square rate.

### 2.2 The Robust Logitboost Algorithm

**Algorithm 2** *Robust logitboost*, which is very similar to Friedman's *mart* algorithm [7], except for Line 4.

---
1: $F_{i,k} = 0$, $p_{i,k} = \frac{1}{K}$, $k = 0$ to $K - 1$, $i = 1$ to $N$
2: For $m = 1$ to $M$ Do
3:     For $k = 0$ to $K - 1$ Do
4:       $\{R_{j,k,m}\}_{j=1}^{J}$ = $J$-terminal node regression tree from
       $\{r_{i,k} - p_{i,k}, \mathbf{x}_i\}_{i=1}^{N}$, with weights $p_{i,k}(1 - p_{i,k})$ as in (7)
5:       $\beta_{j,k,m} = \frac{K-1}{K} \frac{\sum_{\mathbf{x}_i \in R_{j,k,m}} r_{i,k} - p_{i,k}}{\sum_{\mathbf{x}_i \in R_{j,k,m}} (1 - p_{i,k}) p_{i,k}}$
6:       $F_{i,k} = F_{i,k} + \nu \sum_{j=1}^{J} \beta_{j,k,m} \mathbf{1}_{\mathbf{x}_i \in R_{j,k,m}}$
7:     End
8:     $p_{i,k} = \exp(F_{i,k}) / \sum_{s=0}^{K-1} \exp(F_{i,s})$
9: End

---

Alg. 2 describes *robust logitboost* using the tree-split criterion (7). Note that after trees are constructed, the values of the terminal nodes are computed by

$$\frac{\sum_{node} z_{i,k} w_{i,k}}{\sum_{node} w_{i,k}} = \frac{\sum_{node} r_{i,k} - p_{i,k}}{\sum_{node} p_{i,k}(1 - p_{i,k})}, \tag{8}$$

which explains Line 5 of Alg. 2.

### 2.3 Friedman's Mart Algorithm

Friedman [7] proposed *mart* (*multiple additive regression trees*), a creative combination of gradient descent and Newton's method, by using the first-order information to construct the trees and using both the first- & second-order information to determine the values of the terminal nodes.

Corresponding to (7), the tree-split criterion of *mart* is

$$MartGain(s) = \frac{1}{s}\left[\sum_{i=1}^{s}(r_{i,k}-p_{i,k})\right]^2 \qquad (9)$$
$$+ \frac{1}{N-s}\left[\sum_{i=s+1}^{N}(r_{i,k}-p_{i,k})\right]^2 - \frac{1}{N}\left[\sum_{i=1}^{N}(r_{i,k}-p_{i,k})\right]^2.$$

In Sec. 2.1, plugging in responses $z_{i,k} = r_{i,k} - p_{i,k}$ and weights $w_i = 1$, yields (9).

Once the tree is constructed, Friedman [7] applied a one-step Newton update to obtain the values of the terminal nodes. Interestingly, this one-step Newton update yields exactly the same equation as (8). In other words, (8) is interpreted as weighted average in *logitboost* but it is interpreted as the one-step Newton update in *mart*.

Therefore, the *mart* algorithm is similar to Alg. 2; we only need to change Line 4, by replacing (7) with (9).

In fact, Eq. (8) also provides one more explanation why the tree-split criterion (7) is numerically stable, because (7) is always numerically more stable than (8). The update formula (8) has been successfully used in practice for 10 years since the advent of *mart*.

### 2.4 Experiments on Binary Classification

While we focus on multi-class classification, we also provide some experiments on binary classification in App. A.

## 3 Adaptive Base Class (ABC) Logitboost

Developed by [12], the *abc-boost* algorithm consists of the following two components:

1. Using the widely-used *sum-to-zero* constraint [8, 7, 17, 11, 16, 19, 18] on the loss function, one can formulate boosting algorithms only for $K-1$ classes, by using one class as the **base class**.

2. At each boosting iteration, **adaptively** select the base class according to the training loss (3). [12] suggested an exhaustive search strategy.

*Abc-boost* by itself is not a concrete algorithm. [12] developed *abc-mart* by combining *abc-boost* with *mart*. In this paper, we develop **abc-logitboost**, a new algorithm by combining *abc-boost* with *(robust) logitboost*.

Alg. 3 presents *abc-logitboost*, using the derivatives in (5) and (6) and the same exhaustive search strategy as used by *abc-mart*. Again, *abc-logitboost* differs from *abc-mart* only in the tree-split procedure (Line 5 in Alg. 3).

Compared to Alg. 2, *abc-logitboost* differs from *(robust) logitboost* in that they use different derivatives and *abc-logitboost* needs an additional loop to select the base class at each boosting iteration.

**Algorithm 3** *Abc-logitboost* using the exhaustive search strategy for the base class, as suggested in [12]. The vector $B$ stores the base class numbers.

1: $F_{i,k} = 0$, $p_{i,k} = \frac{1}{K}$, $k = 0$ to $K-1$, $i = 1$ to $N$
2: For $m = 1$ to $M$ Do
3:    For $b = 0$ to $K-1$, Do
4:       For $k = 0$ to $K-1$, $k \neq b$, Do
5:          $\{R_{j,k,m}\}_{j=1}^{J} = J$-terminal node regression tree from
          $\{-(r_{i,b}-p_{i,b})+(r_{i,k}-p_{i,k}),\ \mathbf{x}_i\}_{i=1}^{N}$ with weights
          $p_{i,b}(1-p_{i,b})+p_{i,k}(1-p_{i,k})+2p_{i,b}p_{i,k}$, as in Sec. 2.1.
6:          $\beta_{j,k,m} = \frac{\sum_{\mathbf{x}_i \in R_{j,k,m}} -(r_{i,b}-p_{i,b})+(r_{i,k}-p_{i,k})}{\sum_{\mathbf{x}_i \in R_{j,k,m}} p_{i,b}(1-p_{i,b})+p_{i,k}(1-p_{i,k})+2p_{i,b}p_{i,k}}$
7:          $G_{i,k,b} = F_{i,k} + \nu \sum_{j=1}^{J} \beta_{j,k,m} 1_{\mathbf{x}_i \in R_{j,k,m}}$
8:       End
9:       $G_{i,b,b} = -\sum_{k \neq b} G_{i,k,b}$
10:       $q_{i,k} = \exp(G_{i,k,b})/\sum_{s=0}^{K-1} \exp(G_{i,s,b})$
11:       $L^{(b)} = -\sum_{i=1}^{N}\sum_{k=0}^{K-1} r_{i,k} \log(q_{i,k})$
12:    End
13:    $B(m) = \underset{b}{\text{argmin}}\ L^{(b)}$
14:    $F_{i,k} = G_{i,k,B(m)}$
15:    $p_{i,k} = \exp(F_{i,k})/\sum_{s=0}^{K-1} \exp(F_{i,s})$
16: End

### 3.1 Why Does the Choice of Base Class Matter?

It matters because of the diagonal approximation; that is, fitting a regression tree for each class at each boosting iteration. To see this, we can take a look at the Hessian matrix, for $K = 3$. Using the original logitboost/mart derivatives (4), the determinant of the Hessian matrix is

$$\begin{vmatrix} \frac{\partial^2 L_i}{\partial p_0^2} & \frac{\partial^2 L_i}{\partial p_0 p_1} & \frac{\partial^2 L_i}{\partial p_0 p_2} \\ \frac{\partial^2 L_i}{\partial p_1 p_0} & \frac{\partial^2 L_i}{\partial p_1^2} & \frac{\partial^2 L_i}{\partial p_1 p_2} \\ \frac{\partial^2 L_i}{\partial p_2 p_0} & \frac{\partial^2 L_i}{\partial p_2 p_1} & \frac{\partial^2 L_i}{\partial p_2^2} \end{vmatrix}$$
$$= \begin{vmatrix} p_0(1-p_0) & -p_0 p_1 & -p_0 p_2 \\ -p_1 p_0 & p_1(1-p_1) & -p_1 p_2 \\ -p_2 p_0 & -p_2 p_1 & p_2(1-p_2) \end{vmatrix} = 0$$

as expected, because there are only $K-1$ degrees of freedom. A simple fix is to use the diagonal approximation [8, 7]. In fact, when trees are used as the base learner, it seems one must use the diagonal approximation.

Next, we consider the derivatives (5) and (6). This time, when $K=3$ and $k=0$ is the base class, we only have a 2 by 2 Hessian matrix, whose determinant is

$$\begin{vmatrix} \frac{\partial^2 L_i}{\partial p_1^2} & \frac{\partial^2 L_i}{\partial p_1 p_2} \\ \frac{\partial^2 L_i}{\partial p_2 p_1} & \frac{\partial^2 L_i}{\partial p_2^2} \end{vmatrix} =$$
$$\begin{vmatrix} p_0(1-p_0)+p_1(1-p_1)+2p_0 p_1 & p_0 - p_0^2 + p_0 p_1 + p_0 p_2 - p_1 p_2 \\ p_0 - p_0^2 + p_0 p_1 + p_0 p_2 - p_1 p_2 & p_0(1-p_0)+p_2(1-p_2)+2p_0 p_2 \end{vmatrix}$$
$$= p_0 p_1 + p_0 p_2 + p_1 p_2 - p_0 p_1^2 - p_0 p_2^2 - p_1 p_2^2 - p_2 p_1^2 - p_1 p_0^2 - p_2 p_0^2$$
$$+ 6 p_0 p_1 p_2,$$

which is non-zero and is in fact independent of the choice of the base class (even though we assume $k=0$ as the base in this example). In other words, the choice of the base class would not matter if the full Hessian is used.

However, the choice of the base class will matter because we will have to use diagonal approximation in order to construct trees at each iteration.

## 4 Experiments on Multi-class Classification
### 4.1 Datasets

Table 1 lists the datasets used in our study.

Table 1: Datasets

| dataset | $K$ | # training | # test | # features |
|---|---|---|---|---|
| Covertype290k | 7 | 290506 | 290506 | 54 |
| Covertype145k | 7 | 145253 | 290506 | 54 |
| Poker525k | 10 | 525010 | 500000 | 25 |
| Poker275k | 10 | 275010 | 500000 | 25 |
| Poker150k | 10 | 150010 | 500000 | 25 |
| Poker100k | 10 | 100010 | 500000 | 25 |
| Poker25kT1 | 10 | 25010 | 500000 | 25 |
| Poker25kT2 | 10 | 25010 | 500000 | 25 |
| Mnist10k | 10 | 10000 | 60000 | 784 |
| M-Basic | 10 | 12000 | 50000 | 784 |
| M-Rotate | 10 | 12000 | 50000 | 784 |
| M-Image | 10 | 12000 | 50000 | 784 |
| M-Rand | 10 | 12000 | 50000 | 784 |
| M-RotImg | 10 | 12000 | 50000 | 784 |
| M-Noise1 | 10 | 10000 | 2000 | 784 |
| M-Noise2 | 10 | 10000 | 2000 | 784 |
| M-Noise3 | 10 | 10000 | 2000 | 784 |
| M-Noise4 | 10 | 10000 | 2000 | 784 |
| M-Noise5 | 10 | 10000 | 2000 | 784 |
| M-Noise6 | 10 | 10000 | 2000 | 784 |
| Letter15k | 26 | 15000 | 5000 | 16 |
| Letter4k | 26 | 4000 | 16000 | 16 |
| Letter2k | 26 | 2000 | 18000 | 16 |

**Covertype**  The original UCI *Covertype* dataset is fairly large, with $581012$ samples. To generate *Covertype290k*, we randomly split the original data into halves, one half for training and another half for testing. For *Covertype145k*, we randomly select one half from the training set of *Covertype290k* and still keep the same test set.

**Poker**  The UCI *Poker* dataset originally had 25010 samples for training and 1000000 samples for testing. Since the test set is very large, we randomly divide it equally into two parts (I and II). *Poker25kT1* uses the original training set for training and Part I of the original test set for testing. *Poker25kT2* uses the original training set for training and Part II of the original test set for testing. This way, *Poker25kT1* can use the test set of *Poker25kT2* for validation, and *Poker25kT2* can use the test set of *Poker25kT1* for validation. The two test sets are still very large.

In addition, we enlarge the training set to form *Poker525k*, *Poker275k*, *Poker150k*, *Poker100k*. All four enlarged training sets use the same test set as *Pokere25kT2* (i.e., Part II of the original test set). The training set of *Poker525k* contains the original (25k) training set plus Part I of the original test set. The training set of *Poker275k/Poker150k/Poker100k* contains the original training set plus 250k/125k/75k samples from Part I of the original test set.

**Mnist**  While the original *Mnist* dataset is extremely popular, it is known to be too easy [10]. Originally, *Mnist* used 60000 samples for training and 10000 samples for testing. *Mnist10k* uses the original (10000) test set for training and the original (60000) training set for testing.

**Mnist with Many Variations**
[10] created a variety of difficult datasets by adding background (correlated) noises, background images, rotations, etc, to the original *Mnist* data. We shortened the names of the datasets to be *M-Basic*, *M-Rotate*, *M-Image*, *M-Rand*, *M-RotImg*, and *M-Noise1*, *M-Noise2* to *M-Noise6*.

**Letter**  The UCI *Letter* dataset has in total 20000 samples. In our experiments, *Letter4k* (*Letter2k*) use the last 4000 (2000) samples for training and the rest for testing. The purpose is to demonstrate the performance of the algorithms using only small training sets. We also include *Letter15k*, which is one of the standard partitions, by using 15000 samples for training and 5000 samples for testing.

### 4.2 The Main Goal of Our Experiments

The main goal of our experiments is to demonstrate that

1. *Abc-logitboost* and *abc-mart* outperform *(robust) logitboost* and *mart*, respectively.
2. *(Robust) logitboost* often outperforms *mart*.
3. *Abc-logitboost* often outperforms *abc-mart*.
4. The improvements hold for (almost) all reasonable parameters, not just for a few selected sets of parameters.

The main parameter is $J$, the number of terminal tree nodes. It is often the case that test errors are not very sensitive to the shrinkage parameter $\nu$, provided $\nu \leq 0.1$ [7, 3].

### 4.3 Detailed Experiment Results on *Mnist10k*, *M-Image*, *Letter4k*, and *Letter2k*

For these datasets, we experiment with every combination of $J \in \{4, 6, 8, 10, 12, 14, 16, 18, 20, 24, 30, 40, 50\}$ and $\nu \in \{0.04, 0.06, 0.08, 0.1\}$. We train the four boosting algorithms till the training loss (3) is close to the machine accuracy to exhaust the capacity of the learners, for reliable comparisons, up to $M = 10000$ iterations. We report the test mis-classification errors at the last iterations.

For *Mnist10k*, Table 2 presents the test mis-classification errors, which verifies the consistent improvements of (A) *abc-logitboost* over *(robust) logitboost*, (B) *abc-logitboost* over *abc-mart*, (C) *(robust) logitboost* over *mart*, and (D) *abc-mart* over *mart*. The table also verifies that the performances are not too sensitive to the parameters, especially considering the number of test samples is 60000. In App. B, Table 12 reports the testing $P$-values for every combination of $J$ and $\nu$.

Table 3, 4, 5 present the test mis-classification errors on *M-Image*, *Letter4k*, and *Letter2k*, respectively.

Fig. 1 provides the test errors for all boosting iterations. While we believe this is the most reliable comparison, unfortunately there is no space to present them all.

Table 2: *Mnist10k*. Upper table: The test mis-classification errors of *mart* and ***abc-mart*** (bold numbers). Bottom table: The test errors of *logitboost* and ***abc-logitboost*** (bold numbers)

|  | *mart* | ***abc-mart*** | | |
|---|---|---|---|---|
|  | $\nu = 0.04$ | $\nu = 0.06$ | $\nu = 0.08$ | $\nu = 0.1$ |
| $J = 4$ | 3356 **3060** | 3329 **3019** | 3318 **2855** | 3326 **2794** |
| $J = 6$ | 3185 **2760** | 3093 **2626** | 3129 **2656** | 3217 **2590** |
| $J = 8$ | 3049 **2558** | 3054 **2555** | 3054 **2534** | 3035 **2577** |
| $J = 10$ | 3020 **2547** | 2973 **2521** | 2990 **2520** | 2978 **2506** |
| $J = 12$ | 2927 **2498** | 2917 **2457** | 2945 **2488** | 2907 **2490** |
| $J = 14$ | 2925 **2487** | 2901 **2471** | 2877 **2470** | 2884 **2454** |
| $J = 16$ | 2899 **2478** | 2893 **2452** | 2873 **2465** | 2860 **2451** |
| $J = 18$ | 2857 **2469** | 2880 **2460** | 2870 **2437** | 2855 **2454** |
| $J = 20$ | 2833 **2441** | 2834 **2448** | 2834 **2444** | 2815 **2440** |
| $J = 24$ | 2840 **2447** | 2827 **2431** | 2801 **2427** | 2784 **2455** |
| $J = 30$ | 2826 **2457** | 2822 **2443** | 2828 **2470** | 2807 **2450** |
| $J = 40$ | 2837 **2482** | 2809 **2440** | 2836 **2447** | 2782 **2506** |
| $J = 50$ | 2813 **2502** | 2826 **2459** | 2824 **2469** | 2786 **2499** |
|  | *logitboost* | ***abc-logit*** | | |
|  | $\nu = 0.04$ | $\nu = 0.06$ | $\nu = 0.08$ | $\nu = 0.1$ |
| $J = 4$ | 2936 **2630** | 2970 **2600** | 2980 **2535** | 3017 **2522** |
| $J = 6$ | 2710 **2263** | 2693 **2252** | 2710 **2226** | 2711 **2223** |
| $J = 8$ | 2599 **2159** | 2619 **2138** | 2589 **2120** | 2597 **2143** |
| $J = 10$ | 2553 **2122** | 2527 **2118** | 2516 **2091** | 2500 **2097** |
| $J = 12$ | 2472 **2084** | 2468 **2090** | 2468 **2090** | 2464 **2095** |
| $J = 14$ | 2451 **2083** | 2420 **2094** | 2432 **2063** | 2419 **2050** |
| $J = 16$ | 2424 **2111** | 2437 **2114** | 2393 **2097** | 2395 **2082** |
| $J = 18$ | 2399 **2088** | 2402 **2087** | 2389 **2088** | 2380 **2097** |
| $J = 20$ | 2388 **2128** | 2414 **2112** | 2411 **2095** | 2381 **2102** |
| $J = 24$ | 2442 **2174** | 2415 **2147** | 2417 **2129** | 2419 **2138** |
| $J = 30$ | 2468 **2235** | 2434 **2237** | 2423 **2221** | 2449 **2177** |
| $J = 40$ | 2551 **2310** | 2509 **2284** | 2518 **2257** | 2531 **2260** |
| $J = 50$ | 2612 **2353** | 2622 **2359** | 2579 **2332** | 2570 **2341** |

Table 4: *Letter4k*. Upper table: The test mis-classification errors of *mart* and ***abc-mart*** (bold numbers). Bottom table: The test errors of *logitboost* and ***abc-logitboost*** (bold numbers)

|  | *mart* | ***abc-mart*** | | |
|---|---|---|---|---|
|  | $\nu = 0.04$ | $\nu = 0.06$ | $\nu = 0.08$ | $\nu = 0.1$ |
| $J = 4$ | 1681 **1415** | 1660 **1380** | 1671 **1368** | 1655 **1323** |
| $J = 6$ | 1618 **1320** | 1584 **1288** | 1588 **1266** | 1577 **1240** |
| $J = 8$ | 1531 **1266** | 1522 **1246** | 1516 **1192** | 1521 **1184** |
| $J = 10$ | 1499 **1228** | 1463 **1208** | 1479 **1186** | 1470 **1185** |
| $J = 12$ | 1420 **1213** | 1434 **1186** | 1409 **1170** | 1437 **1162** |
| $J = 14$ | 1410 **1190** | 1388 **1156** | 1377 **1151** | 1396 **1160** |
| $J = 16$ | 1395 **1167** | 1402 **1156** | 1396 **1157** | 1387 **1146** |
| $J = 18$ | 1376 **1164** | 1375 **1139** | 1357 **1127** | 1352 **1152** |
| $J = 20$ | 1386 **1154** | 1397 **1130** | 1371 **1131** | 1370 **1149** |
| $J = 24$ | 1371 **1148** | 1348 **1155** | 1374 **1164** | 1391 **1150** |
| $J = 30$ | 1383 **1174** | 1406 **1174** | 1401 **1177** | 1404 **1209** |
| $J = 40$ | 1458 **1211** | 1455 **1224** | 1441 **1233** | 1454 **1215** |
| $J = 50$ | 1484 **1203** | 1517 **1233** | 1487 **1248** | 1522 **1250** |
|  | *logitboost* | ***abc-logit*** | | |
|  | $\nu = 0.04$ | $\nu = 0.06$ | $\nu = 0.08$ | $\nu = 0.1$ |
| $J = 4$ | 1460 **1296** | 1471 **1241** | 1452 **1202** | 1446 **1208** |
| $J = 6$ | 1390 **1143** | 1394 **1117** | 1382 **1090** | 1374 **1074** |
| $J = 8$ | 1336 **1089** | 1332 **1080** | 1311 **1066** | 1297 **1046** |
| $J = 10$ | 1289 **1062** | 1285 **1067** | 1380 **1034** | 1273 **1049** |
| $J = 12$ | 1251 **1058** | 1247 **1069** | 1261 **1044** | 1243 **1051** |
| $J = 14$ | 1247 **1063** | 1233 **1051** | 1251 **1040** | 1244 **1066** |
| $J = 16$ | 1244 **1074** | 1227 **1068** | 1231 **1047** | 1228 **1046** |
| $J = 18$ | 1243 **1059** | 1250 **1040** | 1234 **1052** | 1220 **1057** |
| $J = 20$ | 1226 **1084** | 1242 **1070** | 1242 **1058** | 1235 **1055** |
| $J = 24$ | 1245 **1079** | 1234 **1059** | 1235 **1058** | 1215 **1073** |
| $J = 30$ | 1232 **1057** | 1247 **1085** | 1229 **1069** | 1230 **1065** |
| $J = 40$ | 1246 **1095** | 1255 **1093** | 1230 **1094** | 1231 **1087** |
| $J = 50$ | 1248 **1100** | 1230 **1108** | 1233 **1120** | 1246 **1136** |

Table 3: *M-Image*. Upper table: The test mis-classification errors of *mart* and ***abc-mart*** (bold numbers). Bottom table: The test of *logitboost* and ***abc-logitboost*** (bold numbers)

|  | *mart* | ***abc-mart*** | | |
|---|---|---|---|---|
|  | $\nu = 0.04$ | $\nu = 0.06$ | $\nu = 0.08$ | $\nu = 0.1$ |
| $J = 4$ | 6536 **5867** | 6511 **5813** | 6496 **5774** | 6449 **5756** |
| $J = 6$ | 6203 **5471** | 6174 **5414** | 6176 **5394** | 6139 **5370** |
| $J = 8$ | 6095 **5320** | 6081 **5251** | 6132 **5141** | 6220 **5181** |
| $J = 10$ | 6076 **5138** | 6104 **5100** | 6154 **5086** | 5332 **4983** |
| $J = 12$ | 6036 **4963** | 6086 **4956** | 6104 **4926** | 6117 **4867** |
| $J = 14$ | 5922 **4885** | 6037 **4866** | 6018 **4789** | 5993 **4839** |
| $J = 16$ | 5914 **4847** | 5937 **4806** | 5940 **4797** | 5883 **4766** |
| $J = 18$ | 5955 **4835** | 5886 **4778** | 5896 **4733** | 5814 **4730** |
| $J = 20$ | 5870 **4749** | 5847 **4722** | 5829 **4707** | 5821 **4727** |
| $J = 24$ | 5816 **4725** | 5766 **4659** | 5785 **4662** | 5752 **4625** |
| $J = 30$ | 5729 **4649** | 5738 **4629** | 5724 **4626** | 5702 **4654** |
| $J = 40$ | 5752 **4619** | 5699 **4636** | 5672 **4597** | 5676 **4660** |
| $J = 50$ | 5760 **4674** | 5731 **4667** | 5723 **4659** | 5725 **4649** |
|  | *logitboost* | ***abc-logit*** | | |
|  | $\nu = 0.04$ | $\nu = 0.06$ | $\nu = 0.08$ | $\nu = 0.1$ |
| $J = 4$ | 5837 **5539** | 5852 **5480** | 5834 **5408** | 5802 **5430** |
| $J = 6$ | 5473 **5076** | 5471 **4925** | 5457 **4950** | 5437 **4919** |
| $J = 8$ | 5294 **4756** | 5285 **4748** | 5193 **4678** | 5187 **4670** |
| $J = 10$ | 5141 **4597** | 5120 **4572** | 5052 **4524** | 5049 **4537** |
| $J = 12$ | 5013 **4432** | 5016 **4455** | 4987 **4416** | 4961 **4389** |
| $J = 14$ | 4914 **4378** | 4922 **4338** | 4906 **4356** | 4895 **4299** |
| $J = 16$ | 4863 **4317** | 4842 **4307** | 4816 **4279** | 4806 **4314** |
| $J = 18$ | 4762 **4301** | 4740 **4255** | 4754 **4230** | 4751 **4287** |
| $J = 20$ | 4714 **4251** | 4734 **4231** | 4693 **4214** | 4703 **4268** |
| $J = 24$ | 4676 **4242** | 4610 **4298** | 4663 **4226** | 4638 **4299** |
| $J = 30$ | 4653 **4351** | 4662 **4307** | 4633 **4311** | 4643 **4286** |
| $J = 40$ | 4713 **4434** | 4724 **4426** | 4760 **4439** | 4768 **4388** |
| $J = 50$ | 4763 **4502** | 4795 **4534** | 4792 **4487** | 4799 **4479** |

Table 5: *Letter2k*. Upper table: The test mis-classification errors of *mart* and ***abc-mart*** (bold numbers). Bottom table: The test errors of *logitboost* and ***abc-logitboost*** (bold numbers)

|  | *mart* | ***abc-mart*** | | |
|---|---|---|---|---|
|  | $\nu = 0.04$ | $\nu = 0.06$ | $\nu = 0.08$ | $\nu = 0.1$ |
| $J = 4$ | 2694 **2512** | 2698 **2470** | 2684 **2419** | 2689 **2435** |
| $J = 6$ | 2683 **2360** | 2664 **2321** | 2640 **2313** | 2629 **2321** |
| $J = 8$ | 2569 **2279** | 2603 **2289** | 2563 **2259** | 2571 **2251** |
| $J = 10$ | 2534 **2242** | 2516 **2215** | 2504 **2210** | 2491 **2185** |
| $J = 12$ | 2503 **2202** | 2516 **2215** | 2473 **2198** | 2492 **2201** |
| $J = 14$ | 2488 **2203** | 2467 **2231** | 2460 **2204** | 2460 **2183** |
| $J = 16$ | 2503 **2219** | 2501 **2219** | 2496 **2235** | 2500 **2205** |
| $J = 18$ | 2494 **2225** | 2497 **2212** | 2472 **2205** | 2439 **2213** |
| $J = 20$ | 2499 **2199** | 2512 **2198** | 2504 **2188** | 2482 **2220** |
| $J = 24$ | 2549 **2200** | 2549 **2191** | 2526 **2218** | 2538 **2248** |
| $J = 30$ | 2579 **2237** | 2566 **2232** | 2574 **2244** | 2574 **2285** |
| $J = 40$ | 2641 **2303** | 2632 **2304** | 2606 **2271** | 2667 **2351** |
| $J = 50$ | 2668 **2382** | 2670 **2362** | 2638 **2413** | 2717 **2367** |
|  | *logitboost* | ***abc-logit*** | | |
|  | $\nu = 0.04$ | $\nu = 0.06$ | $\nu = 0.08$ | $\nu = 0.1$ |
| $J = 4$ | 2629 **2347** | 2582 **2299** | 2580 **2256** | 2572 **2231** |
| $J = 6$ | 2427 **2136** | 2450 **2120** | 2428 **2072** | 2429 **2077** |
| $J = 8$ | 2336 **2080** | 2321 **2049** | 2326 **2035** | 2313 **2037** |
| $J = 10$ | 2316 **2044** | 2306 **2003** | 2314 **2021** | 2307 **2002** |
| $J = 12$ | 2315 **2024** | 2315 **1992** | 2333 **2018** | 2290 **2018** |
| $J = 14$ | 2317 **2022** | 2305 **2004** | 2315 **2006** | 2292 **2030** |
| $J = 16$ | 2302 **2024** | 2299 **2004** | 2286 **2005** | 2262 **1999** |
| $J = 18$ | 2298 **2044** | 2277 **2021** | 2301 **1991** | 2282 **2034** |
| $J = 20$ | 2280 **2049** | 2268 **2021** | 2294 **2024** | 2309 **2034** |
| $J = 24$ | 2299 **2060** | 2326 **2037** | 2285 **2021** | 2267 **2047** |
| $J = 30$ | 2318 **2078** | 2326 **2057** | 2304 **2041** | 2274 **2045** |
| $J = 40$ | 2281 **2121** | 2267 **2079** | 2294 **2090** | 2291 **2110** |
| $J = 50$ | 2247 **2174** | 2299 **2155** | 2267 **2133** | 2278 **2150** |

## 4.4 Experiment Results on *Poker25kT1*, *Poker25kT2*

Recall, to provide a reliable comparison (and validation), we form two datasets *Poker25kT1* and *Poker25kT2* by equally dividing the original test set (1000000 samples) into two parts (I and II). Both use the same training set. *Poker25kT1* uses Part I of the original test set for testing and *Poker25kT2* uses Part II for testing.

Table 6 and Table 7 present the test mis-classification errors, for $J \in \{4, 6, 8, 10, 12, 14, 16, 18, 20\}$, $\nu \in \{0.04, 0.06, 0.08, 0.1\}$, and $M = 10000$ boosting iterations (the machine accuracy is not reached). Comparing these two tables, we can see the corresponding entries are very close to each other, which again verifies that the four boosting algorithms provide reliable results on this dataset. Unlike *Mnist10k*, the test errors, especially using *mart* and *logitboost*, are slightly sensitive to the parameter $J$.

Table 6: **Poker25kT1**. Upper table: The test mis-classification errors of *mart* and **abc-mart** (bold numbers). Bottom table: The test of *logitboost* and **abc-logitboost** (bold numbers).

|        | mart          | abc-mart      |               |               |
|--------|---------------|---------------|---------------|---------------|
|        | $\nu = 0.04$  | $\nu = 0.06$  | $\nu = 0.08$  | $\nu = 0.1$   |
| $J=4$  | 145880 **90323** | 132526 **67417** | 124283 **49403** | 113985 **42126** |
| $J=6$  | 71628 **38017**  | 59046 **36839**  | 48064 **35467**  | 43573 **34879**  |
| $J=8$  | 64090 **39220**  | 53400 **37112**  | 47360 **36407**  | 44131 **35777**  |
| $J=10$ | 60456 **39661**  | 52464 **38547**  | 47203 **36990**  | 46351 **36647**  |
| $J=12$ | 61452 **41362**  | 52697 **39221**  | 46822 **37723**  | 46965 **37345**  |
| $J=14$ | 58348 **42764**  | 56047 **40993**  | 50476 **40155**  | 47935 **37780**  |
| $J=16$ | 63518 **44386**  | 55418 **43360**  | 50612 **41952**  | 49179 **40050**  |
| $J=18$ | 64426 **46463**  | 55708 **45607**  | 54033 **45838**  | 52113 **43040**  |
| $J=20$ | 65528 **49577**  | 59236 **47901**  | 56384 **45725**  | 53506 **44295**  |

|        | logitboost    | abc-logit     |               |               |
|--------|---------------|---------------|---------------|---------------|
|        | $\nu = 0.04$  | $\nu = 0.06$  | $\nu = 0.08$  | $\nu = 0.1$   |
| $J=4$  | 147064 **102905** | 140068 **71450** | 128161 **51226** | 117085 **42140** |
| $J=6$  | 81566 **43156**   | 59324 **39164**  | 51526 **37954**  | 48516 **37546**  |
| $J=8$  | 68278 **46076**   | 56922 **40162**  | 52532 **38422**  | 46789 **37345**  |
| $J=10$ | 63796 **44830**   | 55834 **40754**  | 53262 **40486**  | 47118 **38141**  |
| $J=12$ | 66732 **48412**   | 56867 **44886**  | 51248 **42100**  | 47485 **39798**  |
| $J=14$ | 64263 **52479**   | 55614 **48093**  | 51735 **44688**  | 47806 **43048**  |
| $J=16$ | 67092 **53363**   | 58019 **51308**  | 53746 **47831**  | 51267 **46968**  |
| $J=18$ | 69104 **57147**   | 56514 **55468**  | 55290 **50292**  | 51871 **47986**  |
| $J=20$ | 68899 **62345**   | 61314 **57677**  | 56648 **53696**  | 51608 **49864**  |

Table 7: **Poker25kT2**. The test mis-classification errors.

|        | mart          | abc-mart      |               |               |
|--------|---------------|---------------|---------------|---------------|
|        | $\nu = 0.04$  | $\nu = 0.06$  | $\nu = 0.08$  | $\nu = 0.1$   |
| $J=4$  | 144020 **89608** | 131243 **67071** | 123031 **48855** | 113232 **41688** |
| $J=6$  | 71004 **37567**  | 58487 **36345**  | 47564 **34920**  | 42935 **34326**  |
| $J=8$  | 63452 **38703**  | 52990 **36586**  | 46914 **35836**  | 43647 **35129**  |
| $J=10$ | 60061 **39078**  | 52125 **38025**  | 46912 **36455**  | 45863 **36076**  |
| $J=12$ | 61098 **40834**  | 52296 **38657**  | 46458 **37203**  | 46698 **36781**  |
| $J=14$ | 57924 **42348**  | 55622 **40363**  | 50243 **39613**  | 47619 **37243**  |
| $J=16$ | 63213 **44067**  | 55206 **42973**  | 50322 **41485**  | 48966 **39446**  |
| $J=18$ | 64056 **46050**  | 55461 **45133**  | 53652 **45308**  | 51870 **42485**  |
| $J=20$ | 65215 **49046**  | 58911 **47430**  | 56009 **45390**  | 53213 **43888**  |

|        | logitboost    | abc-logit     |               |               |
|--------|---------------|---------------|---------------|---------------|
|        | $\nu = 0.04$  | $\nu = 0.06$  | $\nu = 0.08$  | $\nu = 0.1$   |
| $J=4$  | 145368 **102014** | 138734 **70886** | 126980 **50783** | 116346 **41551** |
| $J=6$  | 80782 **42699**   | 58769 **38592**  | 51202 **37397**  | 48199 **36914**  |
| $J=8$  | 68065 **45737**   | 56678 **39648**  | 52504 **37935**  | 46600 **36731**  |
| $J=10$ | 63153 **44517**   | 55419 **40286**  | 52835 **40044**  | 46913 **37504**  |
| $J=12$ | 66240 **47948**   | 56619 **44602**  | 50918 **41582**  | 47128 **39378**  |
| $J=14$ | 63763 **52063**   | 55238 **47642**  | 51526 **44296**  | 47545 **42720**  |
| $J=16$ | 66543 **52937**   | 57473 **50842**  | 53287 **47578**  | 51106 **46635**  |
| $J=18$ | 68477 **56803**   | 57070 **55166**  | 54954 **49956**  | 51603 **47707**  |
| $J=20$ | 68311 **61980**   | 61047 **57383**  | 56474 **53364**  | 51242 **49506**  |

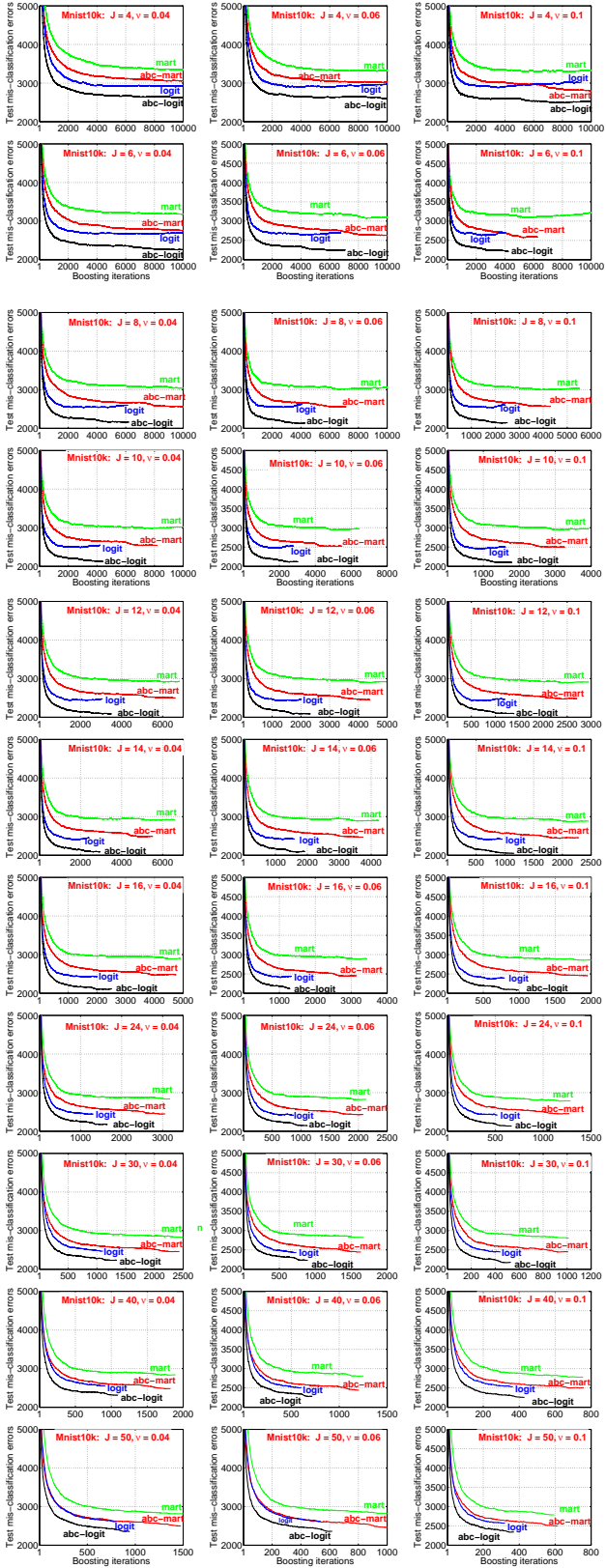

Figure 1: *Mnist10k*. Test mis-classification errors of four boosting algorithms, for shrinkage $\nu = 0.04$ (left), 0.06 (middle), 0.1 (right), and selected $J$ terminal nodes.

## 4.5 Summary of Test Mis-classification Errors

Table 8 summarizes the test mis-classification errors. Since the test errors are not too sensitive to the parameters, for all datasets except *Poker25kT1* and *Poker25kT2*, we simply report the test errors with tree size $J = 20$ and shrinkage $\nu = 0.1$. (More tuning will possibly improve the results.)

For *Poker25kT1* and *Poker25kT2*, as we notice the performance is somewhat sensitive to the parameters, we use each others' test set as the validation set to report the test errors.

For *Covertype290k*, *Poker525k*, *Poker275k*, *Poker150k*, and *Poker100k*, as they are fairly large, we only train $M = 5000$ boosting iterations. For all other datasets, we always train $M = 10000$ iterations or terminate when the training loss (3) is close to the machine accuracy. Since we do not notice obvious over-fitting on these datasets, we simply report the test errors at the last iterations.

Table 8 also includes the results of regular logistic regression. It is interesting that the test errors are all the same (248892) for *Poker525k*, *Poker275k*, *Poker150k*, and *Poker100k* (but the predicted probabilities are different).

Table 8: Summary of test mis-classification errors.

| Dataset | mart | abc-mart | logit | abc-logit | logi. regres. |
|---|---|---|---|---|---|
| Covertype290k | 11350 | 10454 | 10765 | 9727 | 80233 |
| Covertype145k | 15767 | 14665 | 14928 | 13986 | 80314 |
| Poker525k | 7061 | 2424 | 2704 | 1736 | 248892 |
| Poker275k | 15404 | 3679 | 6533 | 2727 | 248892 |
| Poker150k | 22289 | 12340 | 16163 | 5104 | 248892 |
| Poker100k | 27871 | 21293 | 25715 | 13707 | 248892 |
| Poker25kT1 | 43573 | 34879 | 46789 | 37345 | 250110 |
| Poker25kT2 | 42935 | 34326 | 46600 | 36731 | 249056 |
| Mnist10k | 2815 | 2440 | 2381 | 2102 | 13950 |
| M-Basic | 2058 | 1843 | 1723 | 1602 | 10993 |
| M-Rotate | 7674 | 6634 | 6813 | 5959 | 26584 |
| M-Image | 5821 | 4727 | 4703 | 4268 | 19353 |
| M-Rand | 6577 | 5300 | 5020 | 4725 | 18189 |
| M-RotImg | 24912 | 23072 | 22962 | 22343 | 33216 |
| M-Noise1 | 305 | 245 | 267 | 234 | 935 |
| M-Noise2 | 325 | 262 | 270 | 237 | 940 |
| M-Noise3 | 310 | 264 | 277 | 238 | 954 |
| M-Noise4 | 308 | 243 | 256 | 238 | 933 |
| M-Noise5 | 294 | 249 | 242 | 227 | 867 |
| M-Noise6 | 279 | 224 | 226 | 201 | 788 |
| Letter15k | 155 | 125 | 139 | 109 | 1130 |
| Letter4k | 1370 | 1149 | 1252 | 1055 | 3712 |
| Letter2k | 2482 | 2220 | 2309 | 2034 | 4381 |

$P$**-values** Table 9 summarizes four types of $P$-values:

- $P1$: for testing if *abc-mart* has significantly lower ***error rates*** than *mart*.
- $P2$: for testing if *(robust) logitboost* has significantly lower error rates than *mart*.
- $P3$: for testing if *abc-logitboost* has significantly lower error rates than *abc-mart*.
- $P4$: for testing if *abc-logitboost* has significantly lower error rates than *(robust) logitboost*.

The $P$-values are computed using binomial distributions and normal approximations. Recall, if a random variable $z \sim Binomial(N, p)$, then the probability $p$ can be estimated by $\hat{p} = \frac{z}{N}$, and the variance of $\hat{p}$ by $\hat{p}(1 - \hat{p})/N$.

Note that the test sets for *M-Noise1* to *M-Noise6* are very small as [10] did not intend to evaluate the statistical significance on those six datasets. (Private communications.)

Table 9: Summary of test $P$-values.

| Dataset | $P1$ | $P2$ | $P3$ | $P4$ |
|---|---|---|---|---|
| Covertype290k | $3 \times 10^{-10}$ | $3 \times 10^{-5}$ | $9 \times 10^{-8}$ | $8 \times 10^{-14}$ |
| Covertype145k | $4 \times 10^{-11}$ | $4 \times 10^{-7}$ | $2 \times 10^{-5}$ | $7 \times 10^{-9}$ |
| Poker525k | 0 | 0 | 0 | 0 |
| Poker275k | 0 | 0 | 0 | 0 |
| Poker150k | 0 | 0 | 0 | 0 |
| Poker100k | 0 | 0 | 0 | 0 |
| Poker25kT1 | 0 | — | — | 0 |
| Poker25kT2 | 0 | — | — | 0 |
| Mnist10k | $5 \times 10^{-8}$ | $3 \times 10^{-10}$ | $1 \times 10^{-7}$ | $1 \times 10^{-5}$ |
| M-Basic | $2 \times 10^{-4}$ | $1 \times 10^{-8}$ | $1 \times 10^{-5}$ | 0.0164 |
| M-Rotate | 0 | $5 \times 10^{-15}$ | $6 \times 10^{-11}$ | $3 \times 10^{-16}$ |
| M-Image | 0 | 0 | $2 \times 10^{-7}$ | $7 \times 10^{-7}$ |
| M-Rand | 0 | 0 | $7 \times 10^{-10}$ | $8 \times 10^{-4}$ |
| M-RotImg | 0 | 0 | $2 \times 10^{-6}$ | $4 \times 10^{-5}$ |
| M-Noise1 | 0.0029 | 0.0430 | 0.2961 | 0.0574 |
| M-Noise2 | 0.0024 | 0.0072 | 0.1158 | 0.0583 |
| M-Noise3 | 0.0190 | 0.0701 | 0.1073 | 0.0327 |
| M-Noise4 | 0.0014 | 0.0090 | 0.4040 | 0.1935 |
| M-Noise5 | 0.0188 | 0.0079 | 0.1413 | 0.2305 |
| M-Noise6 | 0.0043 | 0.0058 | 0.1189 | 0.1002 |
| Letter15k | 0.0345 | 0.1718 | 0.1449 | 0.0268 |
| Letter4k | $2 \times 10^{-6}$ | 0.008 | 0.019 | $1 \times 10^{-5}$ |
| Letter2k | $2 \times 10^{-5}$ | 0.003 | 0.001 | $4 \times 10^{-6}$ |

These results demonstrate that *abc-logitboost* and *abc-mart* outperform *logitboost* and *mart*, respectively. In addition, except for *Poker25kT1* and *Poker25kT2*, *abc-logitboost* outperforms *abc-mart* and *logitboost* outperforms *mart*.

App. B provides more detailed $P$-values for *Mnsit10k* and *M-Image*, to demonstrate that the improvements hold for a wide range of parameters ($J$ and $\nu$).

## 4.6 Comparisons with SVM and Deep Learning

For *Poker* dataset, SVM could only achieve a test error rate of about $40\%$ (Private communications with C.J. Lin). In comparison, all four algorithms, *mart*, *abc-mart*, *(robust) logitboost*, and *abc-logitboost*, could achieve much smaller error rates (i.e., $< 10\%$) on *Poker25kT1* and *Poker25kT2*.

Fig. 2 provides the comparisons on the six (correlated) noise datasets: *M-Noise1* to *M-Noise6*, with SVM and deep learning based on the results in [10].

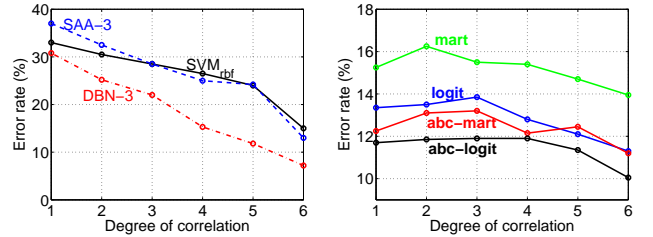

Figure 2: Six datasets: ***M-Noise1*** to ***M-Noise6***. Left panel: Error rates of SVM and deep learning [10]. Right panel: Errors rates of four boosting algorithms. X-axis: degree of correlation from high to low; the values 1 to 6 correspond to the datasets *M-Noise1* to *M-Noise6*.

Table 10: Summary of test error rates of various algorithms on the modified *Mnist* dataset [10].

|            | M-Basic | M-Rotate | M-Image | M-Rand | M-RotImg |
|------------|---------|----------|---------|--------|----------|
| SVM-RBF    | **3.05**% | 11.11%  | 22.61%  | 14.58% | 55.18%   |
| SVM-POLY   | 3.69%   | 15.42%   | 24.01%  | 16.62% | 56.41%   |
| NNET       | 4.69%   | 18.11%   | 27.41%  | 20.04% | 62.16%   |
| DBN-3      | 3.11%   | **10.30**% | 16.31% | **6.73**% | 47.39% |
| SAA-3      | 3.46%   | **10.30**% | 23.00% | 11.28% | 51.93%   |
| DBN-1      | 3.94%   | 14.69%   | 16.15%  | 9.80%  | 52.21%   |
| **mart**   | 4.12%   | 15.35%   | 11.64%  | 13.15% | 49.82%   |
| **abc-mart** | 3.69% | 13.27%   | 9.45%   | 10.60% | 46.14%   |
| **logitboost** | 3.45% | 13.63% | 9.41%   | 10.04% | 45.92%   |
| **abc-logitboost** | 3.20% | 11.92% | **8.54**% | 9.45% | **44.69**% |

Table 10 compares the error rates on *M-Basic*, *M-Rotate*, *M-Image*, *M-Rand*, and *M-RotImg*, with the results in [10].

Fig. 2 and Table 10 illustrate that deep learning algorithms could produce excellent test results on certain datasets (e.g., *M-Rand* and *M-Noise6*). This suggests that there is still sufficient room for improvements in future research.

### 4.7 Test Errors versus Boosting Iterations

Again, we believe the plots for test errors versus boosting iterations could be more reliable than a single number, for comparing boosting algorithms.

Fig. 3 presents the test errors on *Mnist10k*, *M-Rand*, *M-Image*, *Letter15k*, *Letter4k*, and *Letter2k*. Recall we train the algorithms for up to $M = 10000$ iterations unless the training loss (3) is close to the machine accuracy.

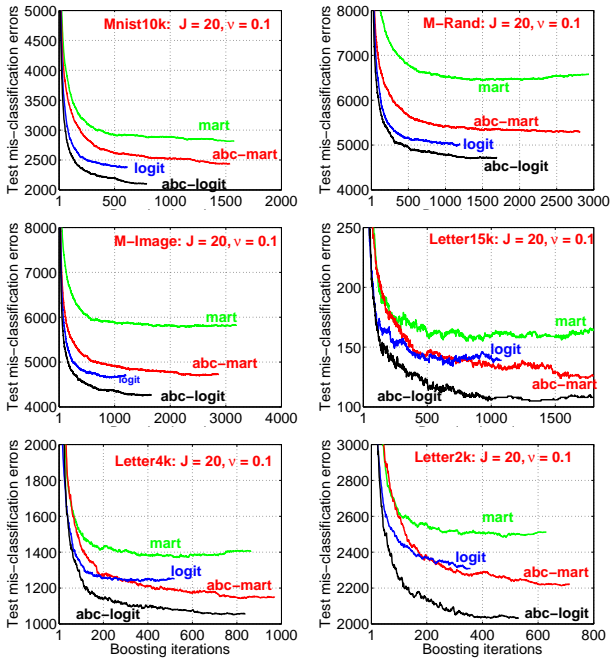

Figure 3: Test mis-classification errors on ***Mnist10k***, ***M-Rand***, ***M-Image***, ***Letter15k***, ***Letter4k***, and ***Letter2k***.

Fig. 4 provides the test mis-classification errors on various datasets from *Covertype* and *Poker*. For these large datasets, we only train $M = 5000$ iterations. (The machine accuracy is not reached.)

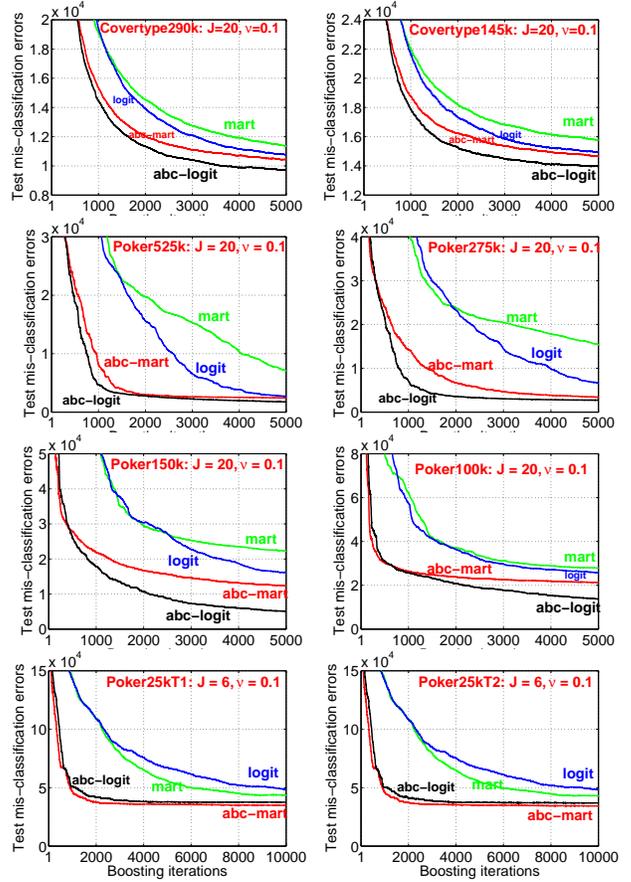

Figure 4: Test mis-classification errors on various datasets of ***Covertype*** and ***Poker***.

### 4.8 Relative Improvements versus Boosting Iterations

For certain applications, it may not be always affordable to use very large models (i.e., many boosting iterations) in the test phrase. Fig. 5 reports the relative improvements (*abc-logitboost* over *(robust) logitboost* and *abc-mart* over *mart*) of the test errors versus boosting iterations.

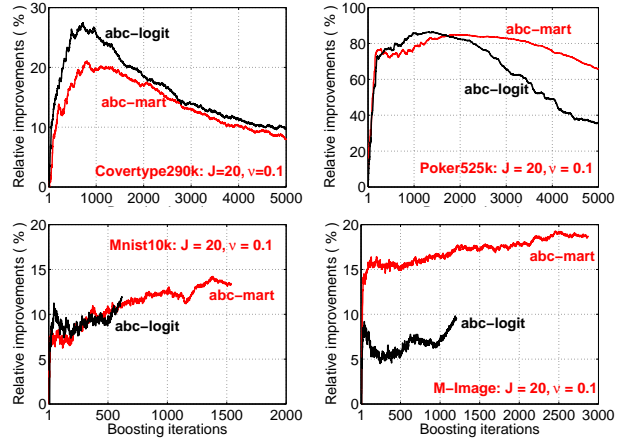

Figure 5: Relative improvements (%) of test errors on ***M-Image***, ***Letter15k***, ***Letter4k***, and ***Letter2k***.

# 5 Conclusion

Classification is a fundamental task in statistics and machine learning. This paper presents *robust logitboost* and *abc-logitboost*, with extensive experiments.

*Robust logitboost* provides the explicit formulation of the tree-split criterion for implementing the influential *logitboost* algorithm. *Abc-logitboost* is developed for multi-class classification, by combining *(robust) logitboost* with *abc-boost*, a new boosting paradigm proposed by [12]. Our extensive experiments demonstrate its superb performance.

We also compare our boosting algorithms with a variety of learning methods including SVM and *deep learning*, using the results in prior publications, e.g., [10]. For certain datasets, *deep learning* obtained adorable performance that our current boosting algorithms could not achieve, suggesting there is still room for improvement in future research.

# Acknowledgement

The author is indebted to Professor Friedman and Professor Hastie for their encouragements and suggestions on this line of work. The author thanks C.J. Lin for the SVM experiments on the *Poker* dataset. The author also thanks L. Bottou and D. Erhan for the explanations on the datasets used in their papers.

The author is partially supported by NSF (DMS-0808864), ONR (N000140910911, Young Investigator Award), and the "Beyond Search - Semantic Computing and Internet Economics" Microsoft 2007 Award.

# A Experiments on Binary Classification

Table 11 lists four datasets for binary classification, to compare *robust logitboost* with *mart*. Fig. 6 reports the results.

Table 11: Datasets for binary classification experiments

| dataset | $K$ | # training | # test | # features |
| --- | --- | --- | --- | --- |
| Mnist2Class | 2 | 60000 | 10000 | 784 |
| IJCNN1 | 2 | 49990 | 91701 | 22 |
| Forest521k | 2 | 521012 | 50000 | 54 |
| Forest100k | 2 | 100000 | 50000 | 54 |

*Forest521k* and *Forest100k* were the two largest datasets in a fairly recent SVM paper [2]. *Mnist2Class* converted the original 10-class MNIST dataset into a binary problem by combining digits from 0 to 4 as one class and 5 to 9 as another class. *IJCNN1* was used in a competition. The winner used SVM (see page 8 at http://www.geocities.com/ijcnn/nnc_ijcnn01.pdf).

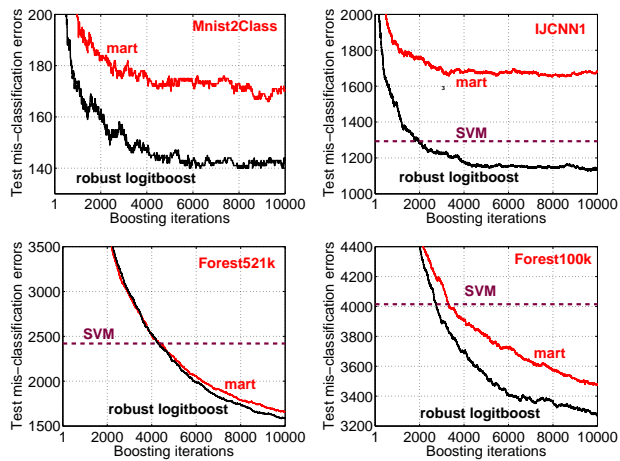

Figure 6: Test mis-classification errors for binary classification. In all experiments, we always use the tree size $J = 20$ and the shrinkage $\nu = 0.1$.

# B  $P$-values for the Experiments on *Mnist10k* and *M-Image*

See Sec. 4.5 for the definitions of P1, P2, P3, and P4. We compute the $P$-values for all combinations of parameters, to show that the improvements are significant not just for one particular set of parameters.

When the author presented this work at various seminars, several researchers were curious about the good performance of our boosting algorithms on the *M-Image* dataset. Thus, we would like to provide more details of the experiments on this dataset.

Table 12: ***Mnist10k***: $P$-values.

| | **P1** | | | |
|---|---|---|---|---|
| | $\nu = 0.04$ | $\nu = 0.06$ | $\nu = 0.08$ | $\nu = 0.1$ |
| $J = 4$ | $7 \times 10^{-5}$ | $3 \times 10^{-5}$ | $7 \times 10^{-10}$ | $1 \times 10^{-12}$ |
| $J = 6$ | $8 \times 10^{-9}$ | $1 \times 10^{-10}$ | $9 \times 10^{-11}$ | 0 |
| $J = 8$ | $9 \times 10^{-12}$ | $4 \times 10^{-12}$ | $5 \times 10^{-13}$ | $2 \times 10^{-10}$ |
| $J = 10$ | $4 \times 10^{-11}$ | $2 \times 10^{-10}$ | $4 \times 10^{-11}$ | $3 \times 10^{-11}$ |
| $J = 12$ | $1 \times 10^{-9}$ | $7 \times 10^{-11}$ | $1 \times 10^{-10}$ | $3 \times 10^{-9}$ |
| $J = 14$ | $6 \times 10^{-10}$ | $1 \times 10^{-9}$ | $6 \times 10^{-9}$ | $9 \times 10^{-10}$ |
| $J = 16$ | $2 \times 10^{-9}$ | $3 \times 10^{-10}$ | $6 \times 10^{-9}$ | $5 \times 10^{-9}$ |
| $J = 18$ | $3 \times 10^{-8}$ | $2 \times 10^{-9}$ | $6 \times 10^{-10}$ | $9 \times 10^{-9}$ |
| $J = 20$ | $2 \times 10^{-8}$ | $3 \times 10^{-8}$ | $2 \times 10^{-8}$ | $6 \times 10^{-8}$ |
| $J = 24$ | $2 \times 10^{-8}$ | $1 \times 10^{-8}$ | $6 \times 10^{-8}$ | $2 \times 10^{-6}$ |
| $J = 30$ | $1 \times 10^{-7}$ | $5 \times 10^{-8}$ | $2 \times 10^{-7}$ | $2 \times 10^{-7}$ |
| $J = 40$ | $3 \times 10^{-7}$ | $1 \times 10^{-7}$ | $2 \times 10^{-8}$ | $5 \times 10^{-5}$ |
| $J = 50$ | $6 \times 10^{-6}$ | $1 \times 10^{-7}$ | $3 \times 10^{-7}$ | $3 \times 10^{-5}$ |
| | **P2** | | | |
| | $\nu = 0.04$ | $\nu = 0.06$ | $\nu = 0.08$ | $\nu = 0.1$ |
| $J = 4$ | $2 \times 10^{-8}$ | $2 \times 10^{-6}$ | $6 \times 10^{-6}$ | $3 \times 10^{-6}$ |
| $J = 6$ | $1 \times 10^{-10}$ | $4 \times 10^{-8}$ | $9 \times 10^{-9}$ | $8 \times 10^{-12}$ |
| $J = 8$ | $4 \times 10^{-10}$ | $2 \times 10^{-9}$ | $1 \times 10^{-10}$ | $1 \times 10^{-9}$ |
| $J = 10$ | $7 \times 10^{-11}$ | $4 \times 10^{-10}$ | $3 \times 10^{-11}$ | $2 \times 10^{-11}$ |
| $J = 12$ | $1 \times 10^{-10}$ | $2 \times 10^{-10}$ | $2 \times 10^{-11}$ | $3 \times 10^{-10}$ |
| $J = 14$ | $2 \times 10^{-11}$ | $8 \times 10^{-12}$ | $2 \times 10^{-10}$ | $3 \times 10^{-11}$ |
| $J = 16$ | $1 \times 10^{-11}$ | $8 \times 10^{-11}$ | $7 \times 10^{-12}$ | $3 \times 10^{-11}$ |
| $J = 18$ | $5 \times 10^{-11}$ | $9 \times 10^{-12}$ | $6 \times 10^{-12}$ | $9 \times 10^{-12}$ |
| $J = 20$ | $2 \times 10^{-10}$ | $2 \times 10^{-9}$ | $1 \times 10^{-9}$ | $4 \times 10^{-10}$ |
| $J = 24$ | $1 \times 10^{-8}$ | $3 \times 10^{-9}$ | $3 \times 10^{-8}$ | $1 \times 10^{-7}$ |
| $J = 30$ | $2 \times 10^{-7}$ | $2 \times 10^{-8}$ | $5 \times 10^{-9}$ | $2 \times 10^{-7}$ |
| $J = 40$ | $3 \times 10^{-5}$ | $1 \times 10^{-5}$ | $4 \times 10^{-6}$ | $2 \times 10^{-4}$ |
| $J = 50$ | $0.0026$ | $0.0023$ | $3 \times 10^{-4}$ | $0.0013$ |
| | **P3** | | | |
| | $\nu = 0.04$ | $\nu = 0.06$ | $\nu = 0.08$ | $\nu = 0.1$ |
| $J = 4$ | $3 \times 10^{-9}$ | $5 \times 10^{-9}$ | $4 \times 10^{-6}$ | $7 \times 10^{-6}$ |
| $J = 6$ | $4 \times 10^{-13}$ | $2 \times 10^{-8}$ | $2 \times 10^{-10}$ | $3 \times 10^{-8}$ |
| $J = 8$ | $2 \times 10^{-9}$ | $3 \times 10^{-10}$ | $3 \times 10^{-10}$ | $6 \times 10^{-11}$ |
| $J = 10$ | $1 \times 10^{-10}$ | $8 \times 10^{-10}$ | $6 \times 10^{-11}$ | $4 \times 10^{-10}$ |
| $J = 12$ | $2 \times 10^{-10}$ | $2 \times 10^{-8}$ | $1 \times 10^{-9}$ | $1 \times 10^{-9}$ |
| $J = 14$ | $5 \times 10^{-10}$ | $6 \times 10^{-9}$ | $4 \times 10^{-10}$ | $4 \times 10^{-10}$ |
| $J = 16$ | $2 \times 10^{-8}$ | $2 \times 10^{-7}$ | $1 \times 10^{-8}$ | $1 \times 10^{-8}$ |
| $J = 18$ | $4 \times 10^{-9}$ | $8 \times 10^{-9}$ | $6 \times 10^{-8}$ | $3 \times 10^{-8}$ |
| $J = 20$ | $1 \times 10^{-6}$ | $2 \times 10^{-7}$ | $6 \times 10^{-8}$ | $2 \times 10^{-7}$ |
| $J = 24$ | $2 \times 10^{-5}$ | $9 \times 10^{-6}$ | $3 \times 10^{-6}$ | $9 \times 10^{-7}$ |
| $J = 30$ | $5 \times 10^{-4}$ | $0.0011$ | $1 \times 10^{-4}$ | $2 \times 10^{-5}$ |
| $J = 40$ | $0.0056$ | $0.0103$ | $0.0024$ | $1 \times 10^{-4}$ |
| $J = 50$ | $0.0145$ | $0.0707$ | $0.0218$ | $0.0102$ |
| | **P4** | | | |
| | $\nu = 0.04$ | $\nu = 0.06$ | $\nu = 0.08$ | $\nu = 0.1$ |
| $J = 4$ | $1 \times 10^{-5}$ | $2 \times 10^{-7}$ | $4 \times 10^{-10}$ | $5 \times 10^{-12}$ |
| $J = 6$ | $5 \times 10^{-11}$ | $7 \times 10^{-11}$ | $1 \times 10^{-12}$ | $6 \times 10^{-13}$ |
| $J = 8$ | $4 \times 10^{-11}$ | $5 \times 10^{-13}$ | $2 \times 10^{-12}$ | $8 \times 10^{-12}$ |
| $J = 10$ | $6 \times 10^{-11}$ | $5 \times 10^{-10}$ | $8 \times 10^{-11}$ | $7 \times 10^{-10}$ |
| $J = 12$ | $2 \times 10^{-9}$ | $6 \times 10^{-9}$ | $6 \times 10^{-9}$ | $1 \times 10^{-8}$ |
| $J = 14$ | $1 \times 10^{-8}$ | $4 \times 10^{-7}$ | $1 \times 10^{-8}$ | $9 \times 10^{-9}$ |
| $J = 16$ | $1 \times 10^{-6}$ | $5 \times 10^{-7}$ | $3 \times 10^{-6}$ | $9 \times 10^{-7}$ |
| $J = 18$ | $1 \times 10^{-6}$ | $8 \times 10^{-7}$ | $2 \times 10^{-6}$ | $8 \times 10^{-6}$ |
| $J = 20$ | $4 \times 10^{-5}$ | $2 \times 10^{-6}$ | $8 \times 10^{-7}$ | $1 \times 10^{-5}$ |
| $J = 24$ | $3 \times 10^{-5}$ | $3 \times 10^{-5}$ | $7 \times 10^{-6}$ | $1 \times 10^{-5}$ |
| $J = 30$ | $3 \times 10^{-4}$ | $0.0016$ | $0.0012$ | $2 \times 10^{-5}$ |
| $J = 40$ | $2 \times 10^{-4}$ | $5 \times 10^{-4}$ | $6 \times 10^{-5}$ | $3 \times 10^{-5}$ |
| $J = 50$ | $9 \times 10^{-5}$ | $7 \times 10^{-5}$ | $2 \times 10^{-4}$ | $4 \times 10^{-4}$ |

Table 13: ***M-Image***: $P$-values.

| | **P1** | | | |
|---|---|---|---|---|
| | $\nu = 0.04$ | $\nu = 0.06$ | $\nu = 0.08$ | $\nu = 0.1$ |
| $J = 4$ | $7 \times 10^{-10}$ | $1 \times 10^{-10}$ | $3 \times 10^{-11}$ | $1 \times 10^{-10}$ |
| $J = 6$ | $5 \times 10^{-12}$ | $6 \times 10^{-13}$ | $1 \times 10^{-13}$ | $3 \times 10^{-13}$ |
| $J = 8$ | $1 \times 10^{-13}$ | 0 | 0 | 0 |
| $J = 10$ | 0 | 0 | 0 | 0 |
| $J = 12$ | 0 | 0 | 0 | 0 |
| $J = 14$ | 0 | 0 | 0 | 0 |
| $J = 16$ | 0 | 0 | 0 | 0 |
| $J = 18$ | 0 | 0 | 0 | 0 |
| $J = 20$ | 0 | 0 | 0 | 0 |
| $J = 24$ | 0 | 0 | 0 | 0 |
| $J = 30$ | 0 | 0 | 0 | 0 |
| $J = 40$ | 0 | 0 | 0 | 0 |
| $J = 50$ | 0 | 0 | 0 | 0 |
| | **P2** | | | |
| | $\nu = 0.04$ | $\nu = 0.06$ | $\nu = 0.08$ | $\nu = 0.1$ |
| $J = 4$ | $1 \times 10^{-10}$ | $1 \times 10^{-9}$ | $1 \times 10^{-9}$ | $2 \times 10^{-9}$ |
| $J = 6$ | $5 \times 10^{-12}$ | $3 \times 10^{-11}$ | $1 \times 10^{-11}$ | $3 \times 10^{-11}$ |
| $J = 8$ | 0 | 0 | 0 | 0 |
| $J = 10$ | 0 | 0 | 0 | 0 |
| $J = 12$ | 0 | 0 | 0 | 0 |
| $J = 14$ | 0 | 0 | 0 | 0 |
| $J = 16$ | 0 | 0 | 0 | 0 |
| $J = 18$ | 0 | 0 | 0 | 0 |
| $J = 20$ | 0 | 0 | 0 | 0 |
| $J = 24$ | 0 | 0 | 0 | 0 |
| $J = 30$ | 0 | 0 | 0 | 0 |
| $J = 40$ | 0 | 0 | 0 | 0 |
| $J = 50$ | 0 | 0 | 0 | 0 |
| | **P3** | | | |
| | $\nu = 0.04$ | $\nu = 0.06$ | $\nu = 0.08$ | $\nu = 0.1$ |
| $J = 4$ | $0.001$ | $8 \times 10^{-4}$ | $0.0003$ | $0.001$ |
| $J = 6$ | $6 \times 10^{-5}$ | $7 \times 10^{-7}$ | $6 \times 10^{-6}$ | $4 \times 10^{-6}$ |
| $J = 8$ | $8 \times 10^{-9}$ | $2 \times 10^{-7}$ | $1 \times 10^{-6}$ | $1 \times 10^{-7}$ |
| $J = 10$ | $2 \times 10^{-8}$ | $3 \times 10^{-8}$ | $4 \times 10^{-9}$ | $2 \times 10^{-6}$ |
| $J = 12$ | $2 \times 10^{-8}$ | $1 \times 10^{-7}$ | $6 \times 10^{-8}$ | $3 \times 10^{-7}$ |
| $J = 14$ | $6 \times 10^{-8}$ | $2 \times 10^{-8}$ | $3 \times 10^{-6}$ | $7 \times 10^{-9}$ |
| $J = 16$ | $1 \times 10^{-8}$ | $8 \times 10^{-8}$ | $2 \times 10^{-8}$ | $9 \times 10^{-7}$ |
| $J = 18$ | $1 \times 10^{-8}$ | $2 \times 10^{-8}$ | $5 \times 10^{-8}$ | $1 \times 10^{-6}$ |
| $J = 20$ | $7 \times 10^{-8}$ | $9 \times 10^{-8}$ | $8 \times 10^{-8}$ | $6 \times 10^{-7}$ |
| $J = 24$ | $1 \times 10^{-7}$ | $6 \times 10^{-5}$ | $2 \times 10^{-6}$ | $3 \times 10^{-5}$ |
| $J = 30$ | $8 \times 10^{-4}$ | $3 \times 10^{-4}$ | $4 \times 10^{-4}$ | $5 \times 10^{-5}$ |
| $J = 40$ | $0.0254$ | $0.0133$ | $0.0475$ | $0.002$ |
| $J = 50$ | $0.0356$ | $0.0818$ | $0.0354$ | $0.0369$ |
| | **P4** | | | |
| | $\nu = 0.04$ | $\nu = 0.06$ | $\nu = 0.08$ | $\nu = 0.1$ |
| $J = 4$ | $0.0025$ | $0.0002$ | $3 \times 10^{-5}$ | $2 \times 10^{-4}$ |
| $J = 6$ | $5 \times 10^{-5}$ | $4 \times 10^{-8}$ | $3 \times 10^{-7}$ | $2 \times 10^{-7}$ |
| $J = 8$ | $8 \times 10^{-9}$ | $2 \times 10^{-7}$ | $1 \times 10^{-6}$ | $1 \times 10^{-7}$ |
| $J = 10$ | $2 \times 10^{-8}$ | $1 \times 10^{-8}$ | $3 \times 10^{-8}$ | $7 \times 10^{-8}$ |
| $J = 12$ | $9 \times 10^{-10}$ | $3 \times 10^{-9}$ | $2 \times 10^{-9}$ | $1 \times 10^{-9}$ |
| $J = 14$ | $1 \times 10^{-8}$ | $5 \times 10^{-10}$ | $5 \times 10^{-10}$ | $2 \times 10^{-10}$ |
| $J = 16$ | $5 \times 10^{-9}$ | $1 \times 10^{-8}$ | $8 \times 10^{-9}$ | $1 \times 10^{-7}$ |
| $J = 18$ | $6 \times 10^{-7}$ | $1 \times 10^{-7}$ | $1 \times 10^{-8}$ | $5 \times 10^{-7}$ |
| $J = 20$ | $5 \times 10^{-7}$ | $5 \times 10^{-8}$ | $2 \times 10^{-7}$ | $2 \times 10^{-6}$ |
| $J = 24$ | $2 \times 10^{-6}$ | $4 \times 10^{-4}$ | $2 \times 10^{-6}$ | $2 \times 10^{-5}$ |
| $J = 30$ | $7 \times 10^{-4}$ | $8 \times 10^{-5}$ | $3 \times 10^{-4}$ | $7 \times 10^{-5}$ |
| $J = 40$ | $0.0017$ | $9 \times 10^{-4}$ | $4 \times 10^{-4}$ | $3 \times 10^{-5}$ |
| $J = 50$ | $0.0032$ | $0.0033$ | $7 \times 10^{-4}$ | $4 \times 10^{-4}$ |